# Performance Analysis of Transformer Based Models (BERT, ALBERT and RoBERTa) in Fake News Detection

Shafna Fitria Nur Azizah
Department of Informatics
Universitas Sebelas Maret
Surakarta, Indonesia
shafitnuraz@student.uns.ac.id

Hasan Dwi Cahyono
Department of Informatics
Universitas Sebelas Maret
Surakarta, Indonesia
hasandc@staff.uns.ac.id

Sari Widya Sihwi
Department of Informatics
Universitas Sebelas Maret
Surakarta, Indonesia
sariwidya@staff.uns.ac.id

Wisnu Widiarto
Department of Informatics
Universitas Sebelas Maret
Surakarta, Indonesia
wisnu.widiarto@staff.uns.ac.id

*Abstract*— Fake news is fake material in a news media format but is not processed properly by news agencies. The fake material can provoke or defame significant entities or individuals or potentially even for the personal interests of the creators, causing problems for society. Distinguishing fake news and real news is challenging due to limited of domain knowledge and time constraints. According to the survey, the top three areas most exposed to hoaxes and misinformation by residents are in Banten, DKI Jakarta and West Java. The model of transformers is referring to an approach in the field of artificial intelligence (AI) in natural language processing utilizing the deep learning architectures. Transformers exercise a powerful attention mechanism to process text in parallel and produce rich and contextual word representations. A previous study indicates a superior performance of a transformer model known as BERT over and above non transformer approach. However, some studies suggest the performance can be improved with the use of improved BERT models known as ALBERT and RoBERTa. However, the modified BERT models are not well explored for detecting fake news in Bahasa Indonesia. In this research, we explore those transformer models and found that ALBERT outperformed other models with 87.6% accuracy, 86.9% precision, 86.9% F1-score, and 174.5 run-time (s/epoch) respectively. Source code available at: https://github.com/Shafna81/fakenewsdetection.git

## I. Introduction

In the era of globalization, internet users have access to a vast amount of information. However, not all information can be considered reliable, as there is a growing concern about the presence of fake news. Fake news refers to false or misleading content presented in a news format, often created without proper verification by news agencies [1]. The dissemination of such false information aims to provoke or defame significant entities or individuals, sometimes driven by personal interests, ultimately causing societal issues [2]. Unfortunately, due to a lack of topic knowledge and time limitations it can be difficult for the typical individual to distinguish between fake news and actual news [3].

Surveys have shown that the regions of Banten, DKI Jakarta, and West Java have the highest exposure to hoaxes and misinformation among their residents [4]. Another survey revealed that a considerable proportion of both rural and urban residents are aware of fake news, with a higher prevalence observed in urban areas [4]. These alarming statistics highlight the severity of the fake news problem in Indonesia.

The Transformer architecture leverages attention mechanisms to process text in parallel and generate rich, contextual word representations [5]. This model focuses on understanding the relationships between words or entities in each text. Transformative models, such as Transformers, have been successfully applied to machine translation, language model, and have contributed to significant progress in NLP [6]. Evaluating the performance of Transformer-based models can measure their ability to identify fake news accurately and efficiently, surpassing architecture based on convolutional layers. This enables developers to create more sophisticated tools and systems for quickly and effectively detecting fake news. Some examples of Transformers models are BERT, ALBERT and RoBERTa.

BERT is an NLP model developed by Google in 2018. Research on detecting fake news using the Transformers BERT model has been previously developed by Jibran Fawaid et al [2] and shows that the BERT model with Transformers Network has the best results with an accuracy of up to 90%. The study used the BERT-Multilingual model, which was trained in 104 languages, including Indonesian, so that the BERT-Multilingual model did not only focus on one type of language. The IndoBERT model is a model specially developed in the case of the Indonesian language [7]. The Transformers A Lite BERT (ALBERT) model is an extension of the BERT model by using the Transformers architecture to pre-train text data but this model has fewer parameters than BERT [8]. ALBERT allows use on devices with limited resources or with faster processing times. Another Transformers model is the RoBERTa model developed by Facebook AI Research (FAIR) in 2019. This model has a similar structure to BERT, but RoBERTa uses more sophisticated training techniques and requires greater data to achieve better performance in various natural language processing tasks [9].

In previous research, the BERT model [6], ALBERT [8], and RoBERTa [9], were each implemented in several NLP cases using English datasets, while this research is a form of replication of Jibran Fawaid's research [2] which discusses the analysis of the detection of fake news with Indonesian language datasets, so researchers want to implement Deep Learning models with models based on Transformers and then see the performance. Fake news detection methods are needed to deal with the problem of fake news effectively and accurately on social networks.

## II. RELATED WORK

Based on [2], Indonesia, a country with the fourth-largest population, is still fighting the fake news. Employing the Transformers Network model (CNN, Bi-LSTM, Hybrid CNN-BiLSTM, and BERT) to identify and address the issue of fake news in Indonesia, the BERT model using the Transformers Network produced the highest results, with up to 90% accuracy. Research on ensemble models for classifying idioms and literal text was done by S. Abarna et al. [9] utilizing the BERT, and RoBERTa models of deep learning.

During the training process, RoBERTa was able to acquire an accuracy level that was 10% greater than BERT. Liu Y et al [9] introduces the RoBERTa pre-training model that is a development of the Bidirectional Encoder Representations from Transformers (BERT) model. The primary objective of RoBERTa is to improve the performance and security of BERT, as well as reduce reliance on specialized data cleaning techniques. RoBERTa achieved better results on GLUE, RACE, and SQuAD. These results highlighted the importance of previously neglected design choices.

Zhenzhong Lan et al [8] introduces a pre-trained ALBERT model that is lighter and more efficient than BERT, but with equivalent performance. ALBERT's pre-trained model has been tested on a variety of NLP tasks such as text understanding, classification, and sentiment analysis. The test results showed that ALBERT can be state-of-the-art on these tasks with fewer parameters and faster processing and has an advantage in adaptation to the task with less data handling compared to BERT.

In a study to identify fake news using machine learning, Suryawanshi A. et al. [10] found that as social media has grown in popularity, more individuals are getting their news from it than from traditional news sources. False information has also been spread using social media, which is harmful to both individual users and society at large. Since social media is significantly better than other online news media in predicting fake news, they concentrated on the detection of false news online media news as well as website verification.

In this study, the research has related to the issue of fake news in Indonesia [2], the models used are the Transformers-based Deep Learning model BERT [6], ALBERT [8] and RoBERTa [9].

## III. MODELS AND METHODS

### A. Deep Learning

Deep learning has been applied in many different applications, including image and video analysis, text processing, natural language translation, autonomous cars, facial and speech recognition, and many more. Deep learning needs a lot of complicated data to train on, and using a GPU or other fast processing device can speed up the training process. However, Deep Learning has the potential to deliver more precise and thorough findings in a range of applications that call for intricate modeling and complicated data processing. In many NLP jobs and other application sectors, deep learning algorithms using Transformer-based networks are being developed continuously. BERT, ALBERT, and RoBERTa are a few examples of Deep Learning algorithms on Transformers-based networks.

### B. Transformers

The Transformers model is a particular Deep Learning architecture that refers to a method for natural language processing in the field of artificial intelligence (AI). Transformers process text in parallel using a potent attention mechanism that results in rich and contextual word representations [5]. This model investigates the connections between textual terms or things. The encoder-decoder structure is included in many competitive neuronal sequence transduction models. A recursive representation sequence ($z = (z_1,..., z_n)$) is created by the encoder from the input symbol representation sequence ($x_1,..., x_n$). The decoder then generates the output sequence of symbols ($y_1,..., y_m$) starting with z [5]. The encoder converts a stream of symbols from the input into a continuous representation. The decoder then creates an output sequence one symbol at a time using the continuous representation provided by the encoder.

Auto-regressive models, which use the encoder-decoder structure, produce the next symbol using the input from the preceding one. To the left and right of Figure 1, respectively, are depicted the Transformers encoder and decoder. Both the encoder and decoder employ completely connected layers that are dots (pointwise) and layers (self-attention). The encoder is made up of N = 6 identical layers. The feed-forward network and the multi-head self-attention mechanism make up the two sub-layers of each layer. After layer normalization, any leftover connections from each sub-layer are cut off. The output of each sub-layer is (LayerNorm(x + Sublayer(x)), where Sub-layers (x) are the individual functions of the sub-layers. The output from all model sub-layers and the embedding layer, which has the size dmodel = 512, enable this residual link.

The N = 6-layer stack used in the decoder is also same. The decoder features an additional third layer that performs multi-head attention on the output of the stack encoder, despite the fact that each encoder layer has two sub-layers. Similar to an encoder, this layer uses residual connections around each of the sub-layers, followed by layer normalization. We also alter the self-attention sub-layers in the decoder stack to stop one position from attending to the position after it.

In order to follow this architecture, the Transformers model uses a self-attention stack and entirely connected layers for the encoder and decoder. Residual connection and layer normalization promote information flow. The attention function determines how many values are at risk based on how close together the key and query are, mapping key-value pairs and queries to output. Transformer use in programs like machine translation and language modeling has considerably advanced NLP [6].

*C. BERT*

Google created the NLP model known as Bidirectional Encoder Representations from Transformers (BERT) in 2018. Using bidirectional training, the BERT model in Transformer then mixes the context from the left and right layers [6]. The BERT model uses the Transformer architecture to study the representation of words in the context of sentences and documents simultaneously. BERT is an unsupervised trained model, meaning it is trained using big data that has no labels so that the model can learn patterns in the data automatically. BERT's ability to understand the context of human language has resulted in major advances in a variety of natural language tasks, including natural language comprehension, natural language processing, and more.

As shown in Figure 2, BERT employs a novel pre-training and fine-tuning strategy. The BERT model is trained on unlabeled large data during the pre-training stage to get a broad knowledge of the language. The model is then adjusted for particular jobs with less data. All parameters are adjusted during fine-tuning.

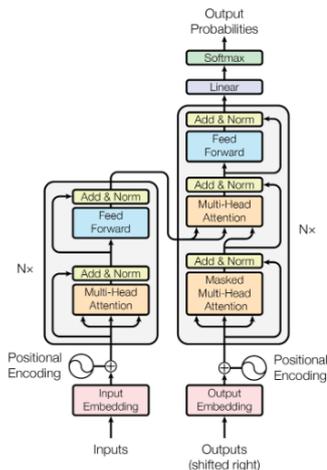

**Fig. 1**. Transformers Model Architecture [5].

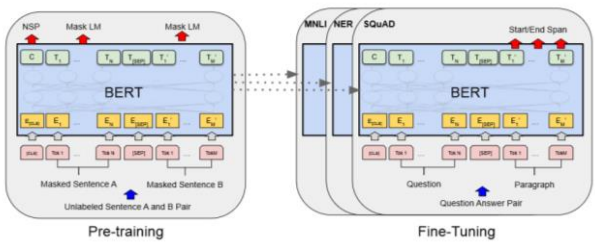

**Figure 2.** Architecture of BERT [6].

Each occurrence of the input is preceded by a unique symbol called [CLS], and [SEP] is a special separator token (such as a separate question and answer). With this method, BERT is able to apply the knowledge it gained during the pre-training phase to certain jobs. BERT is regarded as the state-of-the-art in NLP in part because of this method.

BERT uses the Transformer architecture, which has Attention as its main goal. Attention determines the primary focus or sequence context; as a result, the encoder will also add important keywords and words while translating across languages. Attention is another function that is used to map queries, track key-value pairings, and produce data in the form of vectors [5]. The attention equation can be seen in Equation (1).

$$\textbf{\textit{attention}} \; ( \; Q, \, K, \, V) = \textbf{SoftMax}(\frac{QK^T}{\sqrt{d_k}})V \quad (1)$$

where:
- $Q$ is matrix that constructs the query (contains a vector of each word),
- $K$ is *key*,
- $V$ is *value* itself.
- $d_k$ is dimension of key vector $K$.
- $\frac{QK^T}{\sqrt{d_k}}$ is the step for calculating *attention* weight, which is the result of *the dot product* between $Q$ and $K$ divided by the square root of $d_k$.
- *SoftMax* = one of the activations in deep learning that is useful for attention weights that lead to between 0 and 1 (a kind of probability).

*D. RoBERTa*

*Robustly optimized BERT approach* (RoBERTa) is one of the Natural Language Processing ( *NLP* ) models developed by *Facebook AI Research* (FAIR) in 2019. RoBERTa itself is a development of the BERT model which has proven successful in various language processing tasks. experience. RoBERTa is similar in structure to BERT, but uses more sophisticated training techniques and larger data to achieve better performance in natural language processing tasks, such as natural language comprehension, sentiment analysis, and coarse language detection. RoBERTa has become one of the leading NLP models today and is widely used in various applications that require better and more accurate natural language processing so that even though RoBERTa is not an algorithm directly, it is built using the specific training algorithm mentioned above, and used to process data and produce certain outputs [9]. Therefore, RoBERTa can be categorized as one of the implementations of Deep Learning algorithms in the field of natural language processing.

Figure 3. describes the embedding layers of the RoBERTa model, namely token embedding, and position embedding. One of the main differences between RoBERTa and BERT is the pre-training technique used. BERT is trained using pre-training techniques Masked Language Modeling (MLM), where some words in a sentence are scrambled and the model must predict the missing words. Meanwhile, RoBERTa uses more optimal MLM pre-training techniques and pre-training techniques Next Sentence Prediction (NSP) removed. The RoBERTa pre-training technique also involves a longer training duration and a larger dataset, resulting in a model that is superior in natural language processing. In addition, RoBERTa is also trained using more sophisticated and more diverse data augmentation techniques, including changing

sentence order and token randomization. This helps RoBERTa better understand context and the relationships between words in sentences. Although RoBERTa outperforms BERT in some natural language processing tasks, the two models are similar in architecture and basic principles.

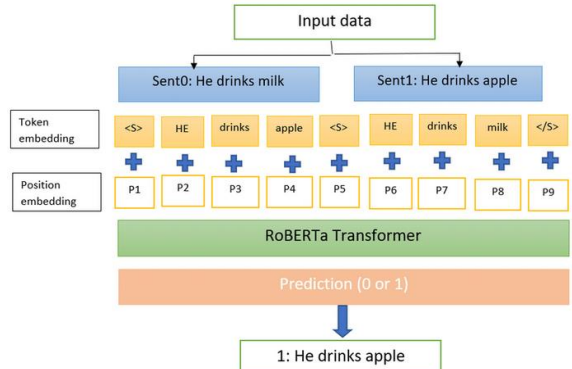

**Fig. 3.** The detail of Embedding Layer in RoBERTa [10].

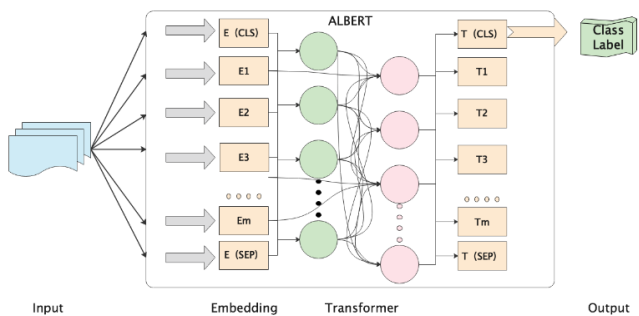

**Fig. 4.** Architecture of ALBERT [3].

*E. ALBERT*

ALBERT or A Lite BERT (Bidirectional Encoder Representations from Transformers) is a pre-trained model developed by Google in 2020. This model is a lighter variant of BERT which was originally introduced in 2018. ALBERT uses the *Transformer architecture* to *pre- train training* on text data. However, this model has a lower number of parameters than BERT, allowing its use in devices with limited resources or with lower processing speeds [11].

ALBERT employs two parameter reduction techniques to eliminate the main constraints for scaling pre-trained models. The first technique uses factorized embedding parameterization, which divides the huge vocabulary insertion matrix into two smaller matrices to separate the sizes of the hidden layer and the embedded vocabulary. Because of this, raising the hidden size is made simpler without considerably growing the vocabulary embedding parameter. The second approach entails sharing parameters between layers. The parameter won't rise with tissue depth using this strategy [8].

ALBERT architecture described in Figure 4. among others:

*a) Embedding Layers*
In the first layer, the words in the sentences will be converted into vector representations in a certain dimensional space through tokenization and embedding processes.

*b) Transformer Encoder Layers*
The text will be processed through a series of Transformer encoder layers. Each layer consists of several sub-layers such as self-attention, feedforward neural network and residual connections and normalization.

*c) Pooling Layer*
After passing through multiple-transformer encoder layers, the output representation of each token in a sequence will be combined using pooling techniques such as mean-pooling or max-pooling to produce a single vector representation.

*d) Output Layers*
The single Vector Representation resulting from the pooling is then input to the fully connected layer to make class/label predictions.

By considerably reducing the number of parameters for BERT while retaining sufficient performance, both methods improve parameter efficiency. With 18x (eighteen times) fewer parameters and a training duration of around 1.7x (one point seven times) faster, the ALBERT configuration, which is comparable to BERT-large, is faster. Techniques for parameter reduction serve as a type of regularization that stabilizes training and promotes generalization. We also added a self-contained loss for sentence-order prediction (SOP) to further boost ALBERT's performance. The main SOP, which concentrated on sentence coherence, was created to address the shortcomings of the next sentence prediction loss (NSP) suggested in the original BERT.

*F. Preprocessing*

Preprocessing data in datasets such as cleaning data from characters or symbols that are not important, removing sentences or words that are not meaningful, and tokenizing text. This preprocessing aims to make data easier to process by the algorithm. Data pre-processing is an important step that involves manipulating data before it is executed, to increase efficiency.

Stopwords are words that are commonly eliminated during the text processing process without changing the content of the phrase because they are thought to have no significance or a very general meaning. Stopwords include phrases like "dan", "yang", "atau" etc. The NLTK directory has a package named corpus, which contains these words that have already been gathered and saved. It may be set up just inside the Python environment. A set of Python-based tools and programs for symbolic and statistical natural language processing (NLP) of English is known as the Natural Language Toolkit (NLTK). The text is broken up into words, and each word is tested to see if it is on the NLTK list of stop words before the finish of the sentence is removed. A word is removed if it appears in the corpus collection [12]. The stopwords section removes non-alphabetical characters such as punctuation marks and numbers.

Tokenization is the process of assigning a numerical value to each word. Text is analyzed to eliminate certain terms before being tokenized in order to use it in predictive modeling. The process of transforming words into integers or floating-point numbers for use as input in machine learning algorithms is known as feature extraction (or vectorization). The discipline of text analysis makes substantial use of machine learning techniques. However, because machine learning methods require fixed-size numeric feature vectors rather than raw text files, raw data, or symbol sequences, most algorithms cannot directly receive them [12].

*G. Hyperparameter*

Hyperparameters are parameters whose values are determined before model training and are not changed during training. Hyperparameters affect model behavior and performance, as well as regulate the training process and model power in generalizing data that has never been seen before [13]. In general, the set hyperparameters consist of batch size, epoch, learning rate, drop out, optimizer, loss function.

## IV. EXPERIMENTS AND EVALUATION

*A. Dataset and Preprocessing*

This dataset was created by integrating three datasets from sources obtained through previous study [2], as in Table 1. A multidimensional sample is a vector that contains all the token frequencies for a single document. A matrix that has one row for each document and one column for each token (such as a word) that appears in the corpus can be used to represent the corpus of documents. The method of vectorization involves turning a group of text documents into a numerical feature vector. Tokenization of BERT-Multilingual and IndoBERT using BertTokenizer, ALBERT using BertTokenizerFast, and RoBERTa using RobertaTokenizer.

*B. Split the Dataset*

The dataset is split into two sections: the training set and the validation set. The validation set is used to test the algorithm's performance while the training set is used to train the algorithm. Data from training and testing are compared 80%: 20%.

*C. Implement the Models*

Create a Deep Learning model using BERT-Multilingual [2], IndoBERT, ALBERT and RoBERTa algorithms using the PyTorch and HuggingFace libraries then selecting the appropriate parameters for the dataset used. The BERT and ALBERT models are taken from the indobenchmark/indobert-base-p1 HuggingFace library, while the RoBERTa model is taken from HuggingFace library cahya/roberta-base-Indonesian-522M.

*D. Training and Testing Models*

Conduct training on the model using a training set. In this study, hyperparameter equalization was carried out in the training and testing of the four models (BERT-Multilingual, IndoBERT, ALBERT, and RoBERTa), as in Table 2. After that, evaluating model performance using a validation set, namely evaluation metrics such as precision, F1 score or accuracy to evaluate model performance. Various evaluation metrics have been applied to gauge how well the algorithm performs in the fake news detection problem. Most current techniques view the issue of fake news as a matter of clarification that determines whether a news piece is fake or not.

**Table 1.** Datasets

| Dataset | Valid | Hoax |
|---|---|---|
| github.com/JibranFawaid/turnbackhoax-dataset | 433 | 683 |
| data.mendeley.com/datasets/p3hfgr5j3m/1 | 372 | 228 |
| github.com/pierobeat/Hoax-NewsClassification | 250 | 250 |

**Table 2.** Model Hyperparameters

| Hyperparameters | Value |
|---|---|
| Batch Size | 16 |
| Epoch | 50 |
| Optimizer | Adam |
| Loss Function | Categorical-crossenthropy |
| Drop Out | 0.5 |
| Learning Rate | 2e-5 |

These approaches include:
- True Positive (TP) : when the predicted piece of fake news is actually annotated as fake news.
- True Negative (TN): when the predicted true news piece is annotated as true news.
- False Negative (FN): when the predicted true news piece is annotated as fake news.
- False Positive (FP): when the predicted piece of fake news is annotated as true news.

By formulating this as a clarification problem, we can define the following metrics (2),(3),(4), and (5):

$$\text{Precision} = \frac{|TP|}{|TP|+|FP|} \quad (2)$$

$$\text{Recall} = \frac{|TP|}{|TP|+|FN|} \quad (3)$$

$$\text{Accuracy} = \frac{|TP|+|TN|}{|TP|+|TN|+|FP|+|FN|} \quad (4)$$

$$\text{F1 score} = 2 \cdot \frac{\text{Precision} \cdot \text{Recall}}{\text{Precision} + \text{Recall}} \quad (5)$$

where:
- Precision is ratio between the number of correctly predicted positive cases and the overall number of positive predicted cases.
- Recall is ratio between the number of correctly predicted positive cases and the total number of positive cases.
- Accuracy is ratio of correct predictions to total predictions in the dataset.
- F1 score is harmonic average of precision and recall.

*E. Results and Discussion*

The research stages when *training* and *testing* are carried out with the Python programming language in the Pytorch *library*. The BERT, ALBERT, and RoBERTa models were trained using Google Collaboration Pro with Python 3 runtime specifications, 25.5 GB of System RAM, T4 GPU hardware accelerator with 15 GB of GPU RAM, 166.8 GB of disk. To maintain consistency, we use a value of random seed 12 for all models.

After setting the hyperparameter, the results are summarized in Table 3 and can be formulated, ALBERT model has the highest accuracy, precision, and F1-Score results when compared to the IndoBERT, BERT-Multilingual, and RoBERTa models. The RoBERTa model runtime performance is 177 seconds/epoch, this result is faster when compared to BERT. Contrarily, ALBERT model also gets a faster runtime. A self-supervised loss for sentence-order prediction (SOP) is able to further boost ALBERT's performance. SOP primarily focuses on inter-sentence coherence and is intended to overcome the inefficiency of the next sentence prediction (NSP) loss suggested in the original BERT [8].

**Table 3.** Performance Comparison and Routine of BERT-Multilingual, IndoBERT, ALBERT, and RoBERTa Models

| Model | Accuracy | Precision | F1-Score | Time (s/epoch) |
|---|---|---|---|---|
| BERT-Multilingual[2] | 90% | 90% | 91% | * |
| BERT-Multilingual(ours) | 78% | 77.7% | 77.7% | 181.54 |
| IndoBERT | 86.6% | 83.4% | 86% | 181.76 |
| RoBERTa | 83.3% | 76.5% | 83.4% | 176.15 |
| **ALBERT** | **87.6%** | **86.9%** | **86.9%** | **174.5** |

*no information on the original paper

## V. CONCLUSION AND FUTURE WORKS

We have followed the steps and datasets based on [2] to detect fake news in Bahasa Indonesia. The outcomes, however, showed that it is challenging to produce an equivalent performance. We hypothesize that the various computational environments may be the root of the performance disparities. Based on the experiment, IndoBERT tokenizer shows stronger performance than BERT-Multilingual in term of accuracy, precision, and F1-score as has been indicated by [7]. In general, ALBERT model shows the most optimal performance results compared to the other models.

We thought about incorporating data augmentation and investigating other hyperparameter tunings can enhance the performance [13]. Our investigation has shown that the inclusion of tokenizers is crucial for enhancing the performance of BERT models. The significance of the tokenizer for RoBERTa and ALBERT has not yet been examined in our study, though. It could be worthwhile to look at other tokenizers for the predefined BERT models.

ACKNOWLEDGMENT

Big thanks for the support of the Faculty of Information Technology and Science Data, the University of Sebelas Maret, Surakarta.